\documentclass[letterpaper]{article} 
\usepackage{myconf2026}  
\usepackage{times}  
\usepackage{helvet}  
\usepackage{courier}  
\usepackage[hyphens]{url}  
\usepackage{graphicx} 
\usepackage{natbib}  
\usepackage{caption} 
\usepackage{booktabs}
\usepackage{multirow}
\usepackage{multirow}
\usepackage{amsmath}
\usepackage{amssymb}
\usepackage{amsthm}
\usepackage{enumitem}
\usepackage{graphicx}
\usepackage{subcaption}
\usepackage[ruled,vlined,linesnumbered]{algorithm2e}
\SetKwInput{Input}{Input}
\SetKwInput{Output}{Output}
\SetKw{Return}{return}
\SetKw{Break}{break}
\SetKw{Continue}{continue}
\usepackage[most]{tcolorbox}
\usepackage{listings}
\usepackage{xcolor}
\newcommand{\myparagraph}[1]{\vspace{0.0mm} \noindent \textbf{#1}.}

\newtheorem{definition}{Definition}

\nocopyright

\title{HyperAgent: Planning and Acting over Tool-Schema Hypergraphs for Tool-Use LLM Agents}
\author {
    Zian Zhai\textsuperscript{\rm 1},
    Xingyu Tan\textsuperscript{\rm 1},
    Gaowang Zou\textsuperscript{\rm 1},
    Xiaoyang Wang\textsuperscript{\rm 1},
    Wenjie Zhang\textsuperscript{\rm 1}
}
\affiliations {
    \textsuperscript{\rm 1}University of New South Wales\\
    zian.zhai@unsw.edu.au, xingyu.tan@unsw.edu.au,  z5445224@ad.unsw.edu.au, xiaoyang.wang1@unsw.edu.au, wenjie.zhang@unsw.edu.au
}
\begin{document}

\maketitle

\begin{abstract}
Large language model (LLM) agents increasingly rely on external tools to complete complex real-world tasks. However, reliable tool-use planning remains challenging due to the limitations of implicit reasoning and the evolving nature of real-world execution environments.
Existing tool-use agents typically rely on LLMs to infer tool compositions from textual descriptions, which can lead to inefficient exploration and unreliable execution in complex tasks.
To address these challenges, we model tool relations at the schema level and construct a directed Tool--Schema Hypergraph, in which tools are represented as hyperedges from their required input-schema nodes to their output-schema nodes.
Furthermore, we propose HyperAgent, a Tool--Schema Hypergraph-guided framework for dynamic planning and execution.
Given a task, HyperAgent first extracts a task-relevant tool context graph and uses it to guide the construction of a schema-aware Task DAG.
During execution, HyperAgent dynamically realizes each subtask by constructing
a state-conditioned tool support graph through deficit-oriented expansion, which identifies unresolved requirements and retrieves supporting producer tools according to the current agent state.
Experiments on AppWorld demonstrate that HyperAgent improves task completion
performance while reducing redundant API calls, LLM interactions, and token
consumption compared with existing agent baselines.
\end{abstract}

\section{Introduction}
With continued improvements in the reasoning capabilities of large language models (LLMs)~\citep{Achiam2023GPT4TR,Yang2024Qwen25TR}, LLM-based autonomous agents are increasingly deployed to perform complex real-world tasks~\citep{Yao2022ReActSR,Kim2025ReflActWD}. 
Equipped with tool-use capabilities, these agents can invoke external APIs, manipulate operating systems, and interact with software services to satisfy user requests.
Such capabilities extend LLMs beyond language generation, allowing agents to take concrete actions and affect their environments~\citep{Yang2023MMREACTPC, Qin2023ToolLLMFL, Patil2023GorillaLL}.
In many practical scenarios, however, fulfilling a user request requires the agent to interpret the task and determine several actions toward the intended goal~\citep{Zhang2025PlanoverGraphTP,Yu2025DynTaskMASAD}.
For tool-using agents, this further entails selecting tools that are aligned with the current objective and executable under the available information, resources, and environment state~\citep{liu2024toolnetconnectinglargelanguage,Lumer2025GraphRF}.

To enhance the task-solving capabilities of LLM agents, numerous approaches have been proposed.
Existing studies have primarily advanced tool-using agents along two complementary directions.
The first focuses on online acting, where the agent repeatedly selects tools, observes its execution result, and determines the next action from the updated state~\citep{Yao2022ReActSR,Kim2025ReflActWD}.
This interaction loop enables the agent to incorporate runtime observations and adapt subsequent actions to execution outcomes.
The second focuses on task planning, where the LLM agent decomposes a user request into intermediate goals or generates a sequence of intended actions before execution~\citep{Sun2023PEARLPL,erdogan2025planandact}.
Such plans provide useful global guidance and help alleviate the myopic behavior of agents that select actions step by step.

Despite this progress, existing approaches have two main limitations.
First, relying on LLMs to infer executable tool chains implicitly from textual tool descriptions is unreliable.
Specifically, for a target operation, the model must identify not only a semantically relevant tool but also the upstream tools required to produce its input parameters.
This dependency reasoning becomes particularly fragile when the agent must process a large volume of tool documentation.
Second, although recent tool-retrieval methods preserve tools directly related to the user request, they may exclude semantically distant yet operationally indispensable prerequisites for the target tools.
Once omitted, these tools can only be recognized during online execution when the chosen tool is found to be non-executable.

These limitations have motivated a growing research interest in graph-based approaches.
For example, at the tool level, ToolNet constrains each tool selection through transitions in a directed tool graph~\citep{liu2024toolnetconnectinglargelanguage}, while Graph RAG--Tool Fusion expands semantically retrieved tools along dependency edges to recover prerequisite tools~\citep{Lumer2025GraphRF}.
At the task level, GNN4TaskPlan employs graph neural networks to select structurally coherent subtasks~\citep{Wu2024CanGL}, and GTool encodes request-specific tool dependencies to guide the generation of complete tool trajectories~\citep{Chen2025GToolGE}.
These methods explicitly model tool relations that need to be inferred implicitly by LLMs, and allow the planner to recover prerequisite tools beyond semantic retrieval.

Nevertheless, existing graph-based approaches remain limited in two critical aspects.
First, current graphs capture only coarse tool dependencies and do not specify which upstream outputs satisfy the inputs of a downstream tool~\citep{Lee2025InNOutAP}.
When multiple tools can produce the same required parameter, this ambiguity makes it difficult to identify the necessary producers and avoid redundant calls.
Second, executable tool composition is state-dependent rather than fixed.
Some required inputs may already be available from the user request or previous executions, thereby changing which upstream tools are still needed.
A statically planned tool path may therefore become redundant or incomplete during execution.

To this end, we first construct a directed Tool--Schema Hypergraph (TSH) to model fine-grained parameter-level relations among tools. Each tool is represented as a directed hyperedge from its input schema nodes to its output schema and effect nodes. 
We further annotate port links between output and input schema nodes to capture schema-level data flow across tools.
Based on the constructed TSH, we propose HyperAgent, a planning and execution framework operating over the TSH. 
Before execution, HyperAgent retrieves a tool context graph that contains potentially useful tools selected according to semantic relevance and hypergraph structure, and uses it to guide the decomposition of the task into a Task DAG whose nodes are connected by identified schema-level dependencies.
During execution, HyperAgent realizes each ready subtask with a Tool Support Subgraph selected according to the current agent state.
The resulting observations update the agent state and are used to refine the remaining Task DAG, so that subsequent tool compositions adapt to the evolving execution context.

In summary, our main contributions are summarized as follows:
\begin{itemize}[leftmargin=*,nosep]
    \item We construct a directed Tool--Schema Hypergraph using a real-world API dataset, which models tool relations as fine-grained schema dependencies.

    \item We propose HyperAgent, a two-stage planning and execution framework that extracts a task-relevant Tool Context Graph and uses it to construct a schema-level Task DAG before execution.

    \item We introduce deficit-oriented support graph expansion to construct state-conditioned Tool Support Subgraphs for ready subtasks, enabling tool compositions to adapt to the dynamical agent state.

    \item Extensive experiments on AppWorld show that HyperAgent improves task completion while reducing redundant API calls, LLM interactions, and token consumption compared with the selected baselines.
\end{itemize}

\section{Related Work}
\myparagraph{Tool-Use Planning}
Tool-use planning enables language agents to select and compose tools into
multi-step action sequences.
Early methods rely on language-model reasoning to guide tool use. ReAct~\citep{Yao2022ReActSR} interleaves reasoning with actions, whereas ART~\citep{paranjape2023artautomaticmultistepreasoning} constructs reasoning programs from retrieved demonstrations.
However, direct trajectory generation provides limited exploration of alternative tool sequences. Search-based methods instead evaluate multiple candidates through depth-first search in ToolLLM~\citep{Qin2023ToolLLMFL} and A*-style search in ToolChain~\citep{zhuang2024toolchain}.
Yet trajectory search still relies on textual descriptions to infer tool dependencies, which may lead to invalid tool selections or infeasible plans.

\myparagraph{Graph-Augmented Agents}
Graph-based methods make tool dependencies explicit by organizing tool
relations and invocation paths into structured graphs
\citep{bei2025graphsmeetaiagents}.
ControlLLM~\citep{liu2024controlllm} searches tool graphs encoding
parameter dependencies, while
ToolNet~\citep{liu2024toolnetconnectinglargelanguage} organizes large
tool collections through directed transitions.
However, tool graphs may be incomplete, irrelevant to a request, or insensitive to the evolving execution trajectory \citep{chen2026gtool,patel-etal-2026-dynamic}. GTool~\citep{chen2026gtool} reduces irrelevant relations through
request-specific construction and addresses incompleteness through
missing-edge prediction, while Dynamic Tool Dependency
Retrieval~\citep{patel-etal-2026-dynamic} updates retrieval using both
the query and current trajectory.
Ordinary graphs represent dependencies through pairwise edges, 
whereas hypergraphs naturally capture higher-order relations involving multiple
entities~\citep{antelmi2023hypergraphsurvey}.
However, their use for executable tool-schema planning under different
agent states remains underexplored.

\section{Preliminaries}

\myparagraph{Tool-Schema Hypergraph (TSH)}
Let $\mathcal H=(\mathcal V,\mathcal E,\mathcal D)$ denote a Tool-Schema Hypergraph, where $\mathcal V$ represents schema and effect nodes, $\mathcal E$ contains tool hyperedges, and $\mathcal D$ contains port-level schema-dependency links between tool hyperedges.
The node set contains input-schema nodes, output-schema nodes, and tool-effect nodes, i.e., 
$\mathcal V=\mathcal V_I\cup \mathcal V_O\cup \mathcal V_F$.
For simplicity, we use $\mathcal V_O$ to denote both output schemas and effect nodes.
Each tool hyperedge $e\in\mathcal E$ is directed from a set of input nodes to a set of output and effect nodes, i.e.,
$
E_e(V_I)=\{V_O\}.$
The TSH contains port-level schema-dependency links $\mathcal D\subseteq \mathcal V_O\times \mathcal V_I,
$ where each dependency $D(v_o,v_i)=w$ indicates that the output schema $v_o$ may support the input schema $v_i$ with weight $w$.
A subhypergraph $\mathcal H_S=(\mathcal V_S,\mathcal E_S,\mathcal D_S)$ is induced by a subset of tool hyperedges $\mathcal E_S\subseteq\mathcal E$,
where $
\mathcal V_S
=
\bigcup_{e\in\mathcal E_S} V(e),
$
and $
\mathcal D_S
=
\{D(v_o,v_i)=w\in\mathcal D
\mid v_o,v_i\in\mathcal V_S\}.
$

\begin{definition}[Neighbor Tools]
Given two tool hyperedges $e_i,e_j\in\mathcal E$, $e_i$ is an upstream neighbor of $e_j$ if at least one of the output schemas of $e_i$ can support at least one of the input schemas of $e_j$ through dependency edges, i.e., $e_i \in \operatorname{Nbr}^{-}(e_j)
\iff
\exists v_o\in O(e_i),\exists v_i\in I(e_j),\exists w>0
\ \text{s.t.}\ 
(v_o,v_i)\in\mathcal D.
$
Symmetrically, $e_j$ is a downstream neighbor of $e_i$, and $
e_j \in \operatorname{Nbr}^{+}(e_i)
\iff
e_i \in \operatorname{Nbr}^{-}(e_j).
$
\end{definition}

\section{Method}
\label{sec:method}

\subsection{HyperGraph Construction}
\myparagraph{Graph Conversion}
We construct the TSH based on In-N-Out~\citep{Lee2025InNOutAP}, an expert-annotated parameter-level API graph. 
In the original graph, API tools and their input and output parameters are represented as nodes.
Directed intra-API links connect input parameters to the corresponding API and the API to its output parameters. 
In addition, directed inter-API schema-dependency links connect an output parameter of one API to an input parameter of another when the returned value can serve as a valid argument for the downstream API. These links are constructed through LLM-based filtering and expert annotation.
Although their tool--schema graph captures schema-level data dependencies across tools, it does not explicitly represent the joint constraints of real-world tool invocation.
Specifically, an API is executable only when all of its required inputs are jointly available and its outputs are jointly produced by the same invocation.
To encode these constraints, we convert the original graph into a directed Tool--Schema Hypergraph.
For each API tool, we represent it as a hyperedge and preserve the original schema nodes $\mathcal{V}$ and inter-tool schema-dependency links $\mathcal{D}$.
\begin{figure*}[t]
  \centering
\includegraphics[width=.9\linewidth]{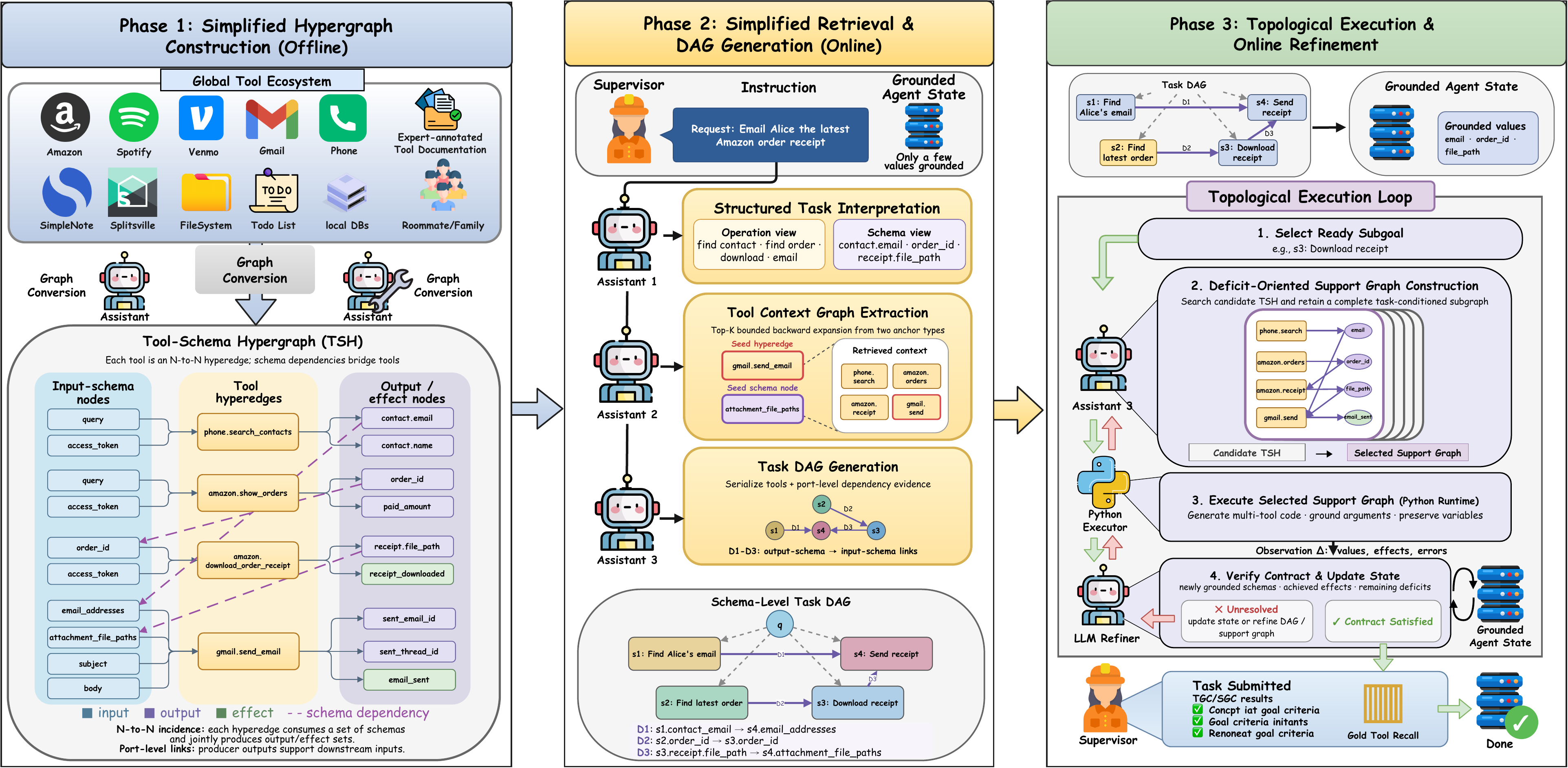}
\caption{The framework of HyperAgent.}
  \label{fig:effeciency}
\end{figure*}

\myparagraph{Graph Refinement}
Some tools do not return reusable parameters but primarily change the environment state, such as sending a message and creating a transaction. Conversely, the execution of a tool may depend on an environment precondition, which is often described in its documentation but is not explicitly presented by input parameters. 
For example, a login tool may establish an authenticated session that is required before invoking a protected API. 
To represent such precondition and effect dependencies, we extract additional effect and condition nodes from API documentation. 
First, we use GPT-4o to extract candidate state changes and execution preconditions from each tool description and align them with the existing input and output schemas. 
Expert annotators then validate the proposed alignments. 
State predicates that are not represented by the original schema nodes are introduced into the node set as additional effect or condition nodes.
We preserve the original port--level schema dependency links in In--N--Out. 
For the newly introduced effect and condition nodes, we enumerate candidate effect--precondition pairs and remove pairs with incompatible domains or entities. Then, we use GPT-4o to score the semantic relevance of the remaining pairs and further filter the irrelevant pairs. 
Subsequently, human annotators verify the filtered candidates and label valid links as strong or weak state dependencies, while other pairs are 
discarded. 
The retained effect--condition links are incorporated into the port--level dependency link set. 
Due to space limitations, we provide implementation details for the refinement procedure and graph statistics in Appendix A.

\subsection{Task-Level Planning}
Given a user request, HyperAgent first extracts a task-relevant Tool Context Graph from the full TSH.
The context graph restricts the planning space to potentially useful tools while retaining the schema-level connections among them.
Conditioned on this context graph, the agent is prompted to decompose the request into a schema--level Task DAG, where each node represents a subtask and each edge specifies
data, effect, or ordering dependencies between subtasks.

\myparagraph{Task Interpretation}
User requests are often expressed at a high level, leaving the intended operations, relevant entities, constraints, and desired environment changes only implicit.
To help the agent locate the relevant region of the TSH, we prompt it to produce a structured task interpretation.
Formally, given a user request $t$, the agent is asked to output a structured representation $d_t =
\left(
d_t^{\mathrm{op}},
d_t^{\mathrm{sch}}
\right)
$, where $d_t^{\mathrm{op}}$ captures the intended operations, and $d_t^{\mathrm{sch}}$ identifies candidate schemas, values, constraints, and desired outcomes expressed or implied by the request.
We use the components of $d_t$ as retrieval signals to anchor relevant tool hyperedges and schema nodes in the TSH.

\myparagraph{Seed Hyperedge Anchoring}
To identify initial task-relevant tools, we retrieve tool hyperedges based on the operation-level interpretation $d_t^{op}$. Specifically, for each hyperedge $e \in \mathcal{E}$, we use its functional description and compute the semantic similarity with $d_t^{op}$, which is denoted as
\begin{equation}
    \operatorname{sim}(e,d_t^{op})
    =
    \cos\left(
        \mathbf{z}(e),
        \mathbf{z}(d_t^{op})
    \right),
    \label{eq:sim}
\end{equation}
where $\mathbf{z}(\cdot)$ denotes the embedding function. 
We select the Top-$K$ hyperedges whose similarity scores exceed a threshold $\theta_e$ to form the seed hyperedge set $\mathcal{E}_{\mathrm{seed}}$.

\myparagraph{Seed Node Anchoring}
We distinguish input and precondition nodes $\mathcal V_I$ from output and effect nodes $\mathcal V_O$.
The schema-level interpretation $d_t^{\mathrm{sch}}$ may contain entities,
constraints, required values, and desired outcomes that correspond directly to these schema nodes.
We therefore use $d_t^{\mathrm{sch}}$ to anchor relevant nodes in the TSH.
For each node $v\in\mathcal V$, we compute the semantic similarity between
its description and $d_t^{\mathrm{sch}}$ using the same embedding-based
similarity function in Equation~\ref{eq:sim}.
We select the top-$K_v$ input-side and output-side nodes whose similarity scores exceed $\theta_v$, yielding $\mathcal V_{\mathrm{seed}}^{I}\subseteq\mathcal V_I$ and $\mathcal V_{\mathrm{seed}}^{O}\subseteq\mathcal V_O$, respectively.
The final seed node set is the union of both sides, i.e., $
\mathcal V_{\mathrm{seed}}
=
\mathcal V_{\mathrm{seed}}^{I}
\cup
\mathcal V_{\mathrm{seed}}^{O}.
$

\myparagraph{Context Graph Extraction}
Starting from the seed hyperedges and nodes, we perform bounded backward search over the TSH to construct a compact, task-relevant context graph.
The expansion reduces the global tool space while retaining the schema--level dependencies connecting the retrieved tools.
We first augment the seed hyperedge set with tools directly associated with the anchored schema nodes. 
For each output-side seed node, tools satisfying $v_o\in O(e)$ are considered as its direct producer candidates.
Similarly, for each input-side seed node, tools satisfying
$v_i\in I(e)$ are considered direct consumer candidates.
Among the producer and consumer hyperedge candidates, we compute the operation-level similarity defined in Equation~\ref{eq:sim} and add the top-$K$ candidates for each seed node to the seed hyperedge set
$\mathcal E_{\mathrm{seed}}$.
We then initialize the backward frontier as
$
\mathcal B_0
=
\mathcal V_{\mathrm{seed}}^{I}
\cup
\{v_i\mid v_i\in I(e),\ e\in\mathcal E_{\mathrm{seed}}\}$.
At each hop $h$, the backward expansion follows dependency links $D$ in reverse towards the output and effect nodes.
Formally, if $v_i\in\mathcal B_h$, $D(v_o,v_i)=w>0$, and $v_o\in O(e_p)$, then
$e_p$ is considered as a producer candidate for $v_i$. 
We rank producer candidates by jointly considering the strength of the
schema-level dependency and their semantic relevance to the operation--level task interpretation:
\begin{equation}
s_{\mathrm{back}}(e,d_t^{\mathrm{op}})
=
w(v_o,v_i)\cdot
\mathrm{cos}\bigl(
\mathbf z(e),
\mathbf z(d_t^{\mathrm{op}})
\bigr).
\end{equation}
For each frontier node, we retain the top-$K$ producer candidates and add them to the retrieved hyperedge set. 
The input nodes of the newly added producers form the next backward frontier $\mathcal B_{h+1}$. 
This expansion continues until the hop budget is exhausted or no new producer is retrieved, inducing the Tool Context Graph.

\begin{definition}[Tool Context Graph]
A Tool Context Graph is the task-relevant subgraph
$\mathcal H_s=(\mathcal V_s,\mathcal E_s,\mathcal D_s)$ of the TSH induced by the seed set,
where $
\mathcal E_s=\mathcal E_{\mathrm{seed}},
\mathcal V_s
=
\bigcup_{e\in\mathcal E_s}
\bigl(I(e)\cup O(e)\bigr),
$
and $
\mathcal D_s=
\{(v_o,v_i)\in\mathcal D
\mid
v_o\in\mathcal V_s,\ v_i\in\mathcal V_s\}.
$
The graph $\mathcal H_s$ provides the LLM planner with a compact context that
preserves both task relevance and the schema-level dependency topology among
the retrieved tools.
\end{definition}

\myparagraph{Task Decomposition}
After constructing the tool context graph $\mathcal H_s$, we serialize it into a compact textual description, where each hyperedge is represented by its API name and description, and each schema node is described by its associated API tool, name, and concise description. 
Moreover, the dependency links are serialized as directed links from the output schemas of producer hyperedges to the input schemas of consumer hyperedges.
We provide the prompt illustration in Appendix B.
Based on this serialized context, the LLM planner is prompted to construct a
Task DAG~\citep{Yang2025AgentNetDE,Dong2024VillagerAgentAG}, where each node represents a subtask and each edge encodes the
identified dependencies among subtasks, including schema-level data and
effect dependencies from the Tool Context Graph.
For each subtask, the planner specifies its local goal, target output schemas
or effects, and a set of candidate terminal tools.
For each dependency between subtasks, the planner identifies relevant schema-dependency links from the Tool Context Graph that connect the output schemas or effects of the upstream subtask to the input schemas required by the downstream subtask.

\begin{definition}[Task DAG]
Given the Tool Context Graph $\mathcal H_s$, a Task DAG is an LLM-generated
directed acyclic graph $\mathcal G=(\mathcal Q,\mathcal L)$.
Each node $q_i\in\mathcal Q$ represents a subtask and is defined as $
q_i=(g_i,\tau_i,\widehat{\mathcal E}_i,\phi_i),
$
where $g_i$ is the local goal,
$\tau_i\subseteq\mathcal V_O$ denotes the target output schemas or effects,
$\widehat{\mathcal E}_i\subseteq\mathcal E_s$ is a small set of candidate
terminal tools, and $\phi_i$ records the execution status of the subtask.
Each edge $\ell_{ij}\in\mathcal L$ represents a data or effect dependency from
$q_i$ to $q_j$.
Such a dependency is grounded in schema-level links from $\mathcal H_s$ that
connect outputs associated with $q_i$ to inputs required by $q_j$.
\end{definition}

\subsection{Task Execution}
\myparagraph{Topological Execution}
Once the initial Task DAG is constructed, the agent executes subtasks following the dependency topology.
At execution step $\ell$, the agent identifies the frontier subtasks whose predecessors have been completed and selects one for execution.
Here, we formally define the agent state.
\begin{definition}[Agent State]
Given the current Task DAG $\mathcal G_\ell=(\mathcal Q_\ell,\mathcal L_\ell)$
before the $\ell$-th execution iteration, the runtime agent state is defined as $
S_\ell=(\mathcal C_\ell,\mathcal R_\ell,\mathcal I_\ell),
$
where $\mathcal C_\ell$ records the execution history, $\mathcal R_\ell$ records the execution status of subtasks, and $\mathcal I_\ell$ stores the runtime schema--value bindings and achieved effects accumulated during execution.
\end{definition}
For the selected subtask, the agent retrieves a tool support subgraph from the TSH for its concrete tool-level realization, which is introduced in the following subsection. 
After executing the selected subtask, the newly grounded schema--value bindings $\Delta\mathcal I_\ell$ and achieved effects update the agent state:
$
S_{\ell+1}=\text{StateUpdate}(S_\ell,\Delta\mathcal I_\ell).
$
The updated state is then used either to execute the next subtask or to refine the remaining Task DAG, yielding $\mathcal G_{\ell+1}$.

\myparagraph{Subtask Verification}
A successful tool call alone is insufficient to verify subtask completion.
After executing the selected tool support graph, HyperAgent determines whether
the execution results satisfy the expected outputs, effects, and constraints
of the current subtask. Specifically, it aligns the execution trace with the
selected support graph and examines the successful API calls, schema--value
bindings, intermediate computation results, and achieved effects. A subtask is
marked as completed only when its required outputs are materialized with valid
schema-level evidence or its target effects are supported by the execution
trace. The verified terminal outputs are then incorporated into $\mathcal I_{\ell+1}$, allowing HyperAgent to determine the initial state of subsequent subtasks. 
Otherwise, HyperAgent retains the observed outputs and effects and refines the current Task DAG. Due to space limitations, the prompt templates and detailed procedure are provided in the Appendix B, C, respectively.

\subsection{Tool--Level Planning}
The Task DAG specifies the subtasks and candidate terminal tools for each
subtask. 
However, HyperAgent does not determine an executable tool composition in advance, as the agent state evolves with newly obtained values
and effects during execution, which may change the required tools.
Instead, HyperAgent dynamically constructs a Tool Support Subgraph for each subtask by identifying unresolved input requirements of candidate terminal tools and expanding producer tools according to the current state.

\myparagraph{Tool--Schema Support}
Before presenting the detailed expansion algorithm, we define Tool-Schema Support to quantify the support strength between
tools and input schemas.

\begin{definition}[Tool-Schema Support]
Given a Tool-Schema Hypergraph
$\mathcal H=(\mathcal V,\mathcal E,\mathcal D)$, let $e\in\mathcal E$ be a tool hyperedge and $r\in\mathcal V_I$ be an input-schema node. 
The Tool-Schema Support from $e$ to $r$ measures the strongest support of the outputs of $e$ to the input schema $r$, and is defined as
\begin{equation}
\rho(e,r)
=
\max_{o\in O(e)}
W(o,r),
\end{equation}
where
\[
W(o,r)=
\begin{cases}
w, & (o,r)\in\mathcal D,\\
0, & \text{otherwise}.
\end{cases}
\]
\end{definition}
If $\rho(e,r)>0$, then $e$ can serve as a producer tool for schema $r$.
For efficient lookup during online execution, we precompute the support between each input-schema node and each tool hyperedge and store the scores in a sparse producer matrix $
\mathbf A_{\mathrm{prod}}
\in \mathbb R^{|\mathcal V_I|\times|\mathcal E|}.$ 
Each row corresponds to an input schema $r$ and contains the support scores of its candidate producer tools, i.e., $
\mathbf A_{\mathrm{prod}}(r,e)=\rho(e,r).$

\myparagraph{Deficit--Oriented Expansion}
For each candidate terminal tool $e_t\in\widehat{\mathcal E}_i$, Deficit--Oriented Expansion (DOE) performs beam search over the TSH to construct a Tool Support Subgraph conditioned on the current agent state $S_\ell$.
Starting from each proposed terminal tool, DOE maintains a deficit set that records input-schema requirements not satisfied by the current agent state or the selected producer
tools.
It iteratively expands candidate subgraphs by adding producer hyperedges to satisfy the deficit set and terminates when a complete support subgraph is found.
For readability, we omit the subtask superscript $q_i$ in this subsection.

\begin{definition}[Deficit Set]
Given an agent state $S_\ell=(\mathcal C_\ell,\mathcal R_\ell,\mathcal I_\ell)$
and a candidate support subgraph $\mathcal G_h=(\mathcal V_h,\mathcal E_h,\mathcal D_h)$, the deficit set $M_{\mathcal G_h}$ contains input-schema nodes required by the selected tools that are neither available in the current agent state nor supported by the output-schema nodes of $\mathcal G_h$ through the retained
dependency links.
\end{definition}

For the candidate terminal tool $e_t$, DOE initializes the candidate subgraph $\mathcal G_0=(\mathcal V_0,\mathcal E_0,\mathcal D_0)$ with the current agent state $S_\ell$, where $
\mathcal V_0=I(e_t),
\mathcal E_0=\{e_t\},$ and $\mathcal D_0=\emptyset.$
The initial deficit set $M_{\mathcal G_0}$ is determined by the input schemas of $e_t$ that are not grounded in the current agent state.
We encode the deficit set as a sparse binary vector $\mathbf m_{\mathcal G_h}\in\{0,1\}^{|\mathcal V_I|\times 1}$, where each dimension indicates whether the corresponding input-schema node belongs to the current deficit set:
\begin{equation}
\mathbf m_{\mathcal G_h}(r)
=
\begin{cases}
1, & r\in M_{\mathcal G_h},\\
0, & r\notin M_{\mathcal G_h},
\end{cases}
\qquad r\in\mathcal V_I.
\label{eq:deficit_mask}
\end{equation}
At each expansion step, DOE selects producer hyperedges that can resolve the current deficits.
Since the output head $O(e)$ of a producer hyperedge may jointly support multiple unresolved input-schema nodes, DOE prioritizes hyperedges whose outputs have greater overlap with the current deficit set $M_{\mathcal G_h}$.
This overlap is defined as
\begin{equation}
\Omega(e,M_{\mathcal G_h})
=
\sum_{r\in M_{\mathcal G_h}} \rho(e,r),
\end{equation}
where $\rho(e,r)$ denotes the Tool-Schema Support defined above.
The top-$K$ producer hyperedges with positive support are selected as expansion directions.
For each selected producer $e_p$, DOE adds the hyperedge together with its input and output schema nodes and retains the dependency links through which its outputs support the nodes in the current deficit set.
The deficit set is then updated after expansion by deleting the resolved deficits and introducing input schemas required by $e_p$ that are not already grounded in the current state or supported by the expanded subgraph.
At each expansion depth, the subgraph candidates are ranked according to the
number of unresolved schema nodes in the deficit set, i.e.,
$|M_{\mathcal G_{h+1}^{(e_p)}}|$, and the top-$B$ candidates with the smallest
deficit sets are retained for the next expansion.
Candidates with empty deficit sets are considered complete support subgraphs
and are no longer expanded.


\myparagraph{Support Graph Selection}
Given a subtask $q_i$, let $\mathbb S_\ell^{q_i}$ denote the set of complete
support subgraphs returned by DOE for the proposed candidate terminal tools in
$\widehat{\mathcal E}_i$.
HyperAgent selects a support subgraph
$\mathcal G_{q_i}\in\mathbb S_\ell^{q_i}$ as the schema-complete tool composition for realizing $q_i$.
If no feasible support subgraph is available or the execution fails, the outcome is incorporated into the agent state and the remaining Task DAG is refined accordingly.

\section{Experiments}
\subsection{Dataset}
We evaluate HyperAgent on the AppWorld dataset~\citep{Trivedi2024AppWorldAC}, a benchmark that evaluates an LLM agent's ability to complete user-directed tasks by interacting with APIs from simulated consumer applications, including email, payment, music, shopping, phone, and file-management services. The agent operates through a stateful Python interpreter and performs tasks of varying difficulty. Task correctness is determined by unit tests that verify whether the requested changes were successfully applied, whether any unintended modifications were introduced, and whether the final answer matches the reference answer when applicable.
The benchmark contains 250 task scenarios, each instantiated with three variants, resulting in 750 tasks in total. These tasks are divided into a training set of 35 scenarios (105 tasks), a development set of 20 scenarios (60 tasks), a normal test set (\textsc{Test-N}) of 56 scenarios (168 tasks), and a challenge test set (\textsc{Test-C}) of 139 scenarios (417 tasks). Compared with \textsc{Test-N}, \textsc{Test-C} requires longer and more complex interaction sequences and may involve applications that are unseen during training. Performance is measured using Task Goal Completion (TGC), which reports the proportion of successfully completed tasks, and Scenario Goal Completion (SGC), which counts a scenario as solved only when all of its task variants are completed successfully.

\begin{figure*}[t]
  \centering
\includegraphics[width=.95\linewidth]{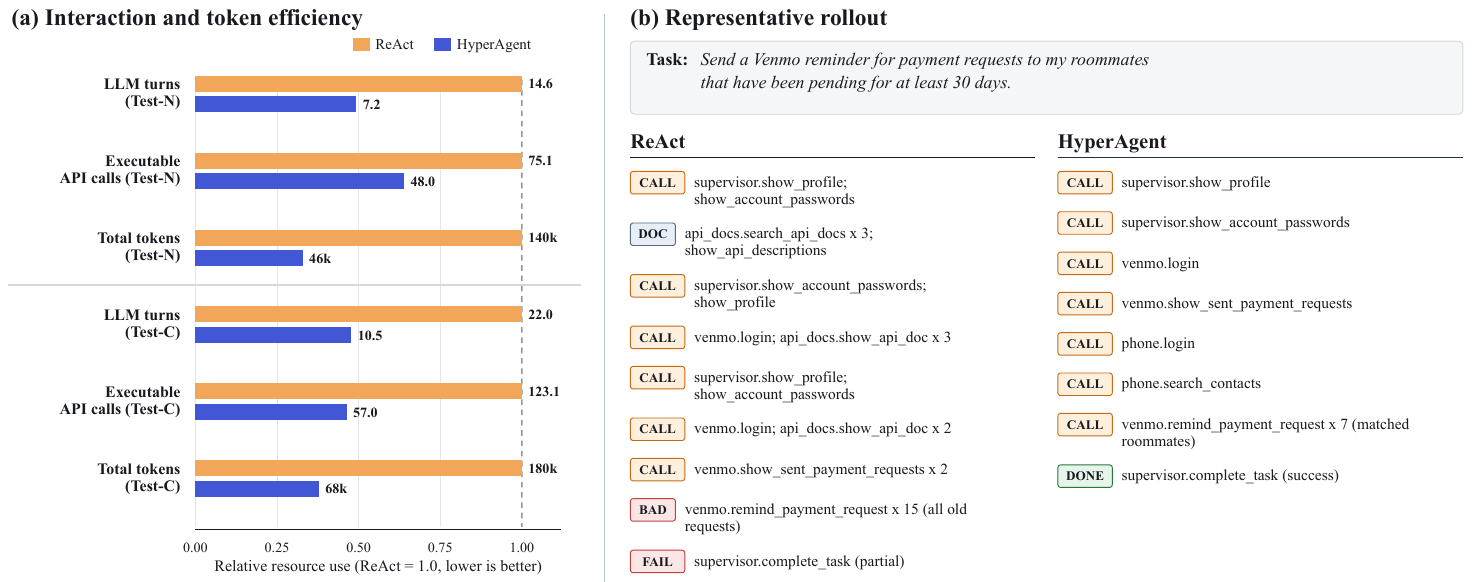}
\caption{Tool call and token usage for HyperAgent.}
  \label{fig:effeciency}
\end{figure*}
\subsection{Baselines}
\myparagraph{Supervised Fine-Tuning (SFT)}
We include three supervised baselines that learn from successful agent trajectories. SFT-GT transforms gold AppWorld traces into ReAct-style interaction trajectories for supervised training. 
RFT~\citep{Yuan2023ScalingRO} samples trajectories from the base agent and fine-tunes the model only on successful rollouts. 
EI~\cite{Anthony2017ThinkingFA} performs RFT iteratively, using the improved policy from each round to collect training trajectories for the next round.

\myparagraph{Direct Preference Optimization (DPO)}
We also compare HyperAgent with preference-based methods. 
DPO-MCTS~\citep{Putta2024AgentQA} uses Monte Carlo tree search to discover alternative action trajectories and constructs preference pairs from their estimated returns. 
DMPO~\citep{Shi2024DirectMP} extends preference optimization to multi-turn interaction by jointly modeling the decisions within preferred and rejected trajectories.

\myparagraph{Reinforcement Learning (RL)}
We include policy-gradient methods that optimize agents directly from environment rewards. 
PPO~\citep{Schulman2017ProximalPO} trains policies using a learned value model for advantage estimation. 
RLOO~\citep{ahmadian2024basicsrevisitingreinforcestyle} estimates trajectory-level advantages by comparing multiple rollouts sampled for the same task. 
GRPO~\citep{Shao2024DeepSeekMathPT} performs clipped group-relative updates using normalized rewards among rollouts of the same task, while LOOP reuses old sampled trajectories through clipped policy updates and assigns credit at trajectory (bandit), turn, or token-level~\citep{Chen2025ReinforcementLF}.

\myparagraph{Non-Fine-Tuning Methods (NFT)}
For NFT methods, we perform four agent scaffolds. 
ReAct~\citep{Yao2022ReActSR} solves tasks through iterative reasoning, execution, and observation. 
Plan and Execution (PlanExec)~\citep{erdogan2025planandact} generates a plan which consists subtasks before ReAct execution, whereas Reflexion (FullCodeRefl)~\citep{Kim2025ReflActWD} explicitly writes and revises execution feedback after each execution. 
Traj (SetBSR+Snippet)~\citep{Gupta2025LeveragingIL} augments the agent with few-shot trajectory demonstrations before the test task and inserts a state-matched snippet into the context before each action decision.

\begin{table}[t]
\centering
\caption{Performance comparison with different paradigms on AppWorld.}
\label{tab:appworld-nestful}
\renewcommand{\arraystretch}{0.95}

\resizebox{\columnwidth}{!}{%
\begin{tabular}{ll|cccc}
\toprule
\multirow{3}{*}{Model}
& \multirow{3}{*}{Prompting}
& \multicolumn{4}{c}{AppWorld}
\\
\cline{3-6}
& & \multicolumn{2}{c}{Test-N}
  & \multicolumn{2}{c}{Test-C}
\\
& & TGC & SGC & TGC & SGC \\
\midrule

\multirow{4}{*}{GPT-4o}
& ReAct        & 48.8 & 32.1 & 30.2 & 13.0  \\
& PlanExec & 44.6 & 23.2 & 19.7 & 7.9\\
& FullCodeRefl      & 33.9 & 26.8 & 19.2 & 12.2 \\
& HyperAgent      & 63.1 & 52.6
& 35.7 & 18.9  \\

\midrule
\multirow{4}{*}{GPT-4turbo}
& ReAct        & 26.8 & 12.5 & 17.5 & 5.8  \\
& PlanExec & 32.7 & 16.1 & 11.0 & 3.6 \\
& FullCodeRefl      & 25.6 & 19.3 & 12.5 & 7.2  \\
& HyperAgent      & 45.3 & 28.6 & 23.1 & 10.1  \\
\midrule

\multirow{4}{*}{Llama-3-70B-Instruct}
& ReAct        & 20.8 & 8.9 & 3.4 & 0  \\
& PlanExec & 8.9 & 1.8 & 2.4 & 0.7  \\
& FullCodeRefl      & 24.4 & 17.9 & 7.0 & 4.3\\
& HyperAgent      & 30.1 & 19.7 & 8.9 & 6.4\\
\bottomrule

\end{tabular}
}
\label{tab:1}
\end{table}

\begin{table}[t]
\centering

\caption{
Performance comparison on AppWorld under different agent
optimization methods.
}
\renewcommand{\arraystretch}{0.95}

\resizebox{\columnwidth}{!}{%
\begin{tabular}{@{}clcccc@{}}
\toprule
\multicolumn{2}{c}{Approach}
& \multicolumn{2}{c}{Test-N}
& \multicolumn{2}{c}{Test-C} \\
\cmidrule(lr){3-4}
\cmidrule(lr){5-6}
&
& TGC & SGC
& TGC & SGC \\
\midrule

\multirow{3}{*}{SFT}
& SFT-GT
& 6.2 & 1.8
& 0.8 & 0.1 \\

& RFT
& 47.9 & 26.4
& 26.4 & 11.4 \\

& EI
& 58.3 & 36.8
& 32.8 & 17.6 \\

\midrule

\multirow{2}{*}{DPO}
& DPO-MCTS
& 57.0 & 31.8
& 31.8 & 13.7 \\

& DMPO
& 59.0 & 36.6
& 36.3 & 13.7 \\

\midrule

\multirow{6}{*}{RL}
& PPO (critic, token)
& 50.8 & 28.9
& 26.4 & 10.5 \\

& RLOO (traj)
& 57.2 & 35.7
& 36.7 & 17.4 \\

& GRPO (token)
& 58.0 & 36.8
& 39.5 & 22.4 \\

& LOOP (bandit)
& 53.3 & 33.6
& 27.7 & 13.0 \\

& LOOP (turn)
& 64.1 & 43.5
& 40.8 & 26.5 \\

& LOOP (token)
& \textbf{71.3} & \textbf{53.6}
& \textbf{45.7} & \textbf{26.6} \\

\midrule

\multirow{3}{*}{NFT}
& ReAct
& 48.8 & 32.1
& 30.2 & 13.0 \\
& PlanExec
& 63.1 & 52.6
& 35.7 & 18.9 \\
& Traj(SsetBSR+Snippet)
& 65.8 & 53.6
& 38.7 & 24.8 \\
& HyperAgent
& 67.1 & 55.9
& 40.2 & 26.1 \\

\bottomrule
\end{tabular}%
}
\label{tab:2}
\end{table}

\subsection{Agent Implementation}
We use GPT-4o as the backbone for all LLM-based components in the main experiments, including hypergraph construction, task interpretation, semantic DAG planning and refinement, support-graph selection, and ReAct-Code execution. Tool context extraction, Deficit-Oriented Expansion, and runtime binding validation are implemented as deterministic modules and do not invoke the LLM unless semantic ambiguity remains. 
Each task is executed in a freshly initialized AppWorld environment with a persistent restricted Python REPL, allowing intermediate variables and tool outputs to be reused across subgoals. 
The prompt templates for the Planner, Refiner, and Executor, together with their decoding configurations, are provided in Appendix~A.
We also vary the LLM backbone, using GPT-4--Turbo and Llama-3--70B--Instruct, to evaluate HyperAgent against other agent frameworks.

\begin{figure*}[t]
  \centering
  \begin{subfigure}[t]{0.39\linewidth}
    \centering
    \includegraphics[width=0.95\linewidth]{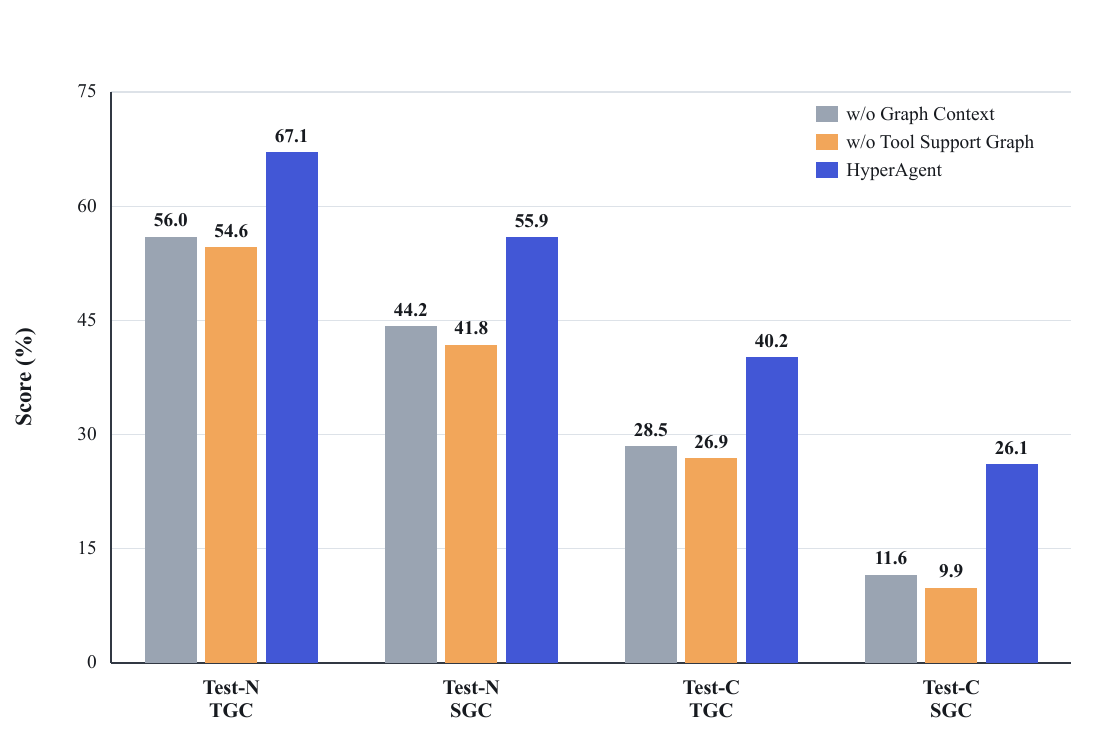}
    \caption{Task Performance.}
    \label{fig:Task}
  \end{subfigure}
  \hfill
  \begin{subfigure}[t]{0.29\linewidth}
    \centering
    \includegraphics[width=0.95\linewidth]{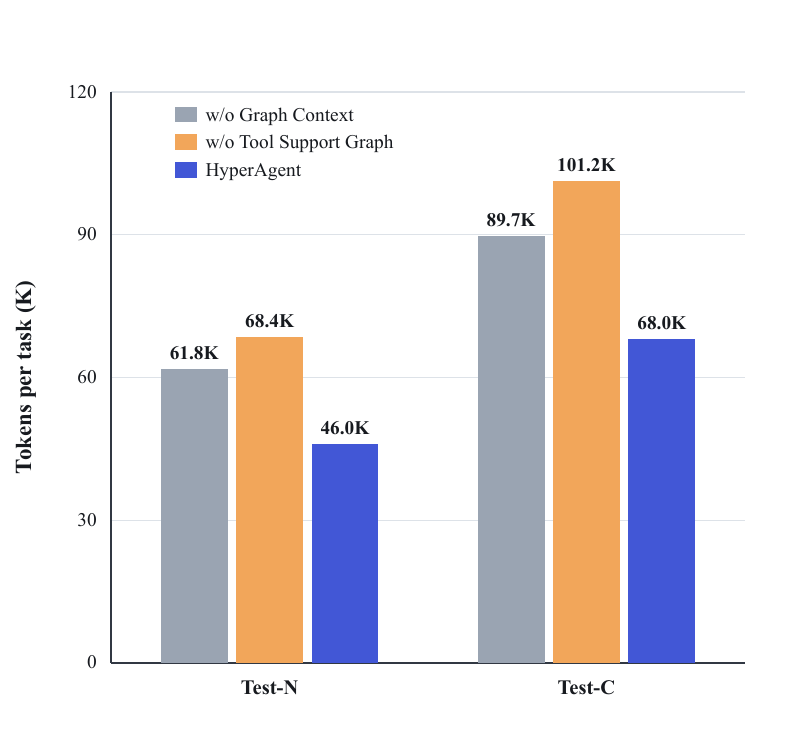}
    \caption{Token Usage.}
    \label{fig:Token}
  \end{subfigure}
  \hfill
  \begin{subfigure}[t]{0.29\linewidth}
    \centering
    \includegraphics[width=0.95\linewidth]{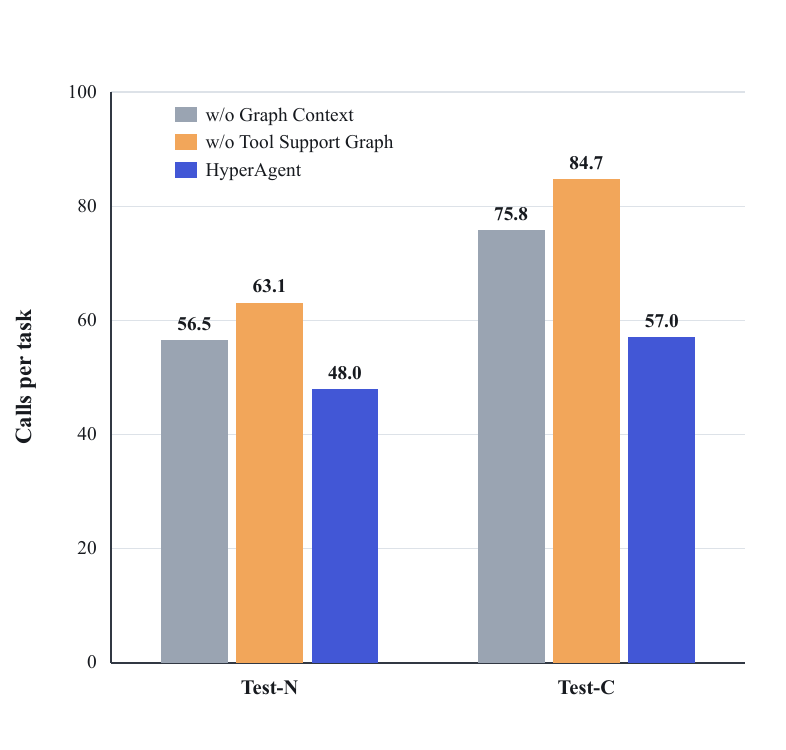}
    \caption{API Calls.}
    \label{fig:Calls}
  \end{subfigure}
  \caption{Ablation study with two variants.}
  \label{fig:ablation}
\end{figure*}
\subsection{Main Results}
\myparagraph{RQ1: Does HyperAgent improve end-to-end task performance}
First, we compare HyperAgent with methods from the SFT, DPO, RL, and NFT paradigms. 
All SFT, DPO, and RL baselines are trained with the Qwen-2.5-32B LLM backbone, whereas the NFT methods are based on GPT-4o.
As shown in Table~\ref{tab:2}, by using a more comprehensive planning context and detailed schema-level Tool Support Subgraphs, HyperAgent outperforms the other NFT methods and is competitive with trained agents.
Second, we compare HyperAgent with different agent scaffolds.
From Table~\ref{tab:1}, we observe that, with schema-level dependencies, HyperAgent consistently outperforms ReAct, PlanExec, and FullCodeRefl across the evaluated LLM backbones.

\myparagraph{RQ2: Does HyperAgent reduce tool-document and total costs}
We compare HyperAgent with ReAct in terms of LLM interaction turns, API calls, and total token consumption on Test-Normal and Test-Challenge. 
As shown in Fig.~\ref{fig:effeciency}, HyperAgent consistently reduces costs on both types of tasks. 
We further present a representative execution rollout (with additional examples in the Appendix D), from which we observe that ReAct repeatedly explores API documentation and retries actions before resolving the necessary constraints, whereas HyperAgent follows dependency-grounded tool sequences to accomplish planned subgoals, such as retrieving roommates' contact information, identifying matching payment requests, and issuing reminders.
With the TSH, HyperAgent constructs prerequisite-complete support paths for individual subgoals, reducing costly online exploration over the action space.

\subsection{Ablation Study}
\begin{figure}[t]
  \centering
  \begin{subfigure}[t]{0.99\linewidth}
    \centering
    \includegraphics[width=0.95\linewidth]{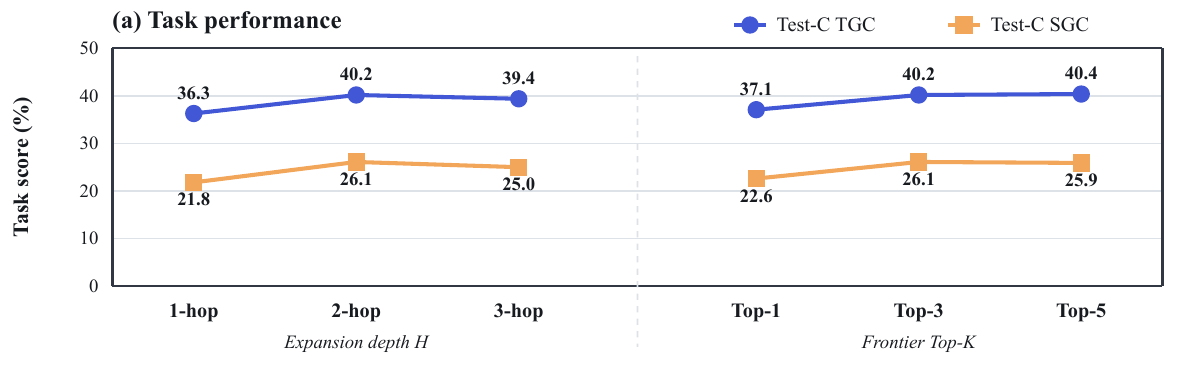}
    \caption{Task Performance.}
  \end{subfigure}
  \hfill
  \begin{subfigure}[t]{0.99\linewidth}
    \centering
    \includegraphics[width=0.95\linewidth]{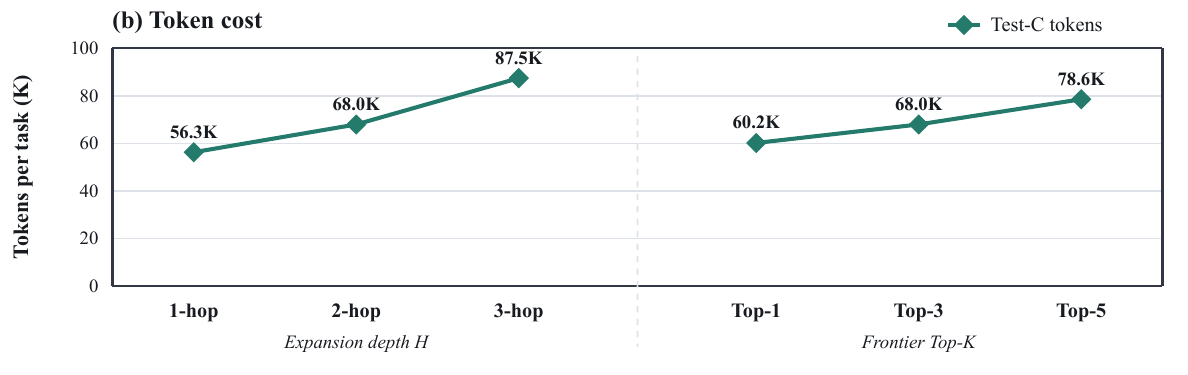}
    \caption{Token Usage.}
  \end{subfigure}
  \caption{Key hyperparameter analysis in HyperAgent.}
  \label{fig:hyperpara}
\end{figure}

\myparagraph{RQ3: Are the main components of HyperAgent essential and effective}
First, we compare HyperAgent with two variants. In \textbf{Variant~1}, we remove the retrieved graph context, including tool-schema dependencies, and retain only simplified tool descriptions during DAG planning and refinement. In \textbf{Variant~2}, we remove the Tool Support Subgraphs for each subtask and use the top-$K$ tools with the highest semantic similarity to the subgoal.
As shown in Figure~\ref{fig:ablation}, HyperAgent achieves higher task performance while using fewer tokens and API calls. 
Removing either component consistently degrades end-to-end task performance on both Test-N and Test-C. 
This trend suggests that graph context and tool support graphs
reduce redundant tool exploration by exposing schema-level
relations across tools. Moreover, HyperAgent can construct more complete execution paths and avoid semantically related but operationally unsuitable tools, leading to fewer API invocations and less exploration.
Second, we study the impact of the hop count in the Tool Context Graph and the top-$K$ support parameter used in support-graph expansion, as shown in Figure~\ref{fig:hyperpara}. 
Increasing the hop count from 1 to 2 and the top-$K$ value from 1 to 3 improves both TGC and SGC by recovering more valid prerequisite paths.
However, further expansion provides negligible gains while substantially increasing token consumption due to additional irrelevant tools and dependencies. 
\subsection{More Experiments}
\myparagraph{RQ4: Does HyperAgent construct a compact tool context that preserves the gold tools}
\begin{figure}[t]
  \centering
\includegraphics[width=.6\linewidth]{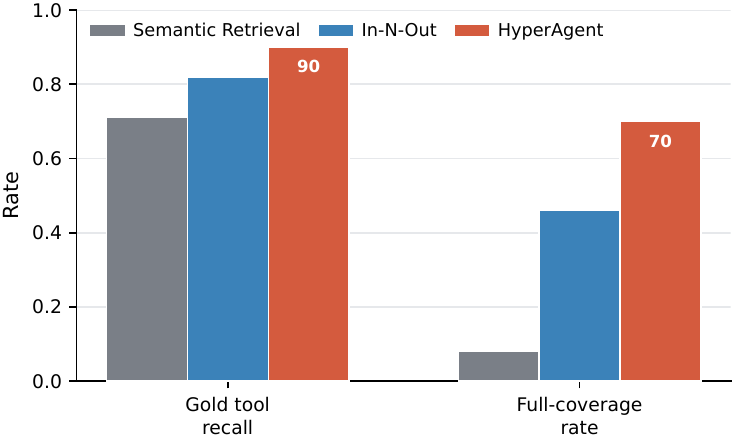}
\caption{Quality evaluation of HyperAgent.}
  \label{fig:rq4}
\end{figure}
To evaluate the effectiveness of the proposed Tool Context Graphs, we extract task-level gold tool sets from the official AppWorld solutions and compare HyperAgent with semantic top-$K$ tool retrieval and In-N-Out graph retrieval under the same context budget of 20 tools.
Based on the results in Figure~\ref{fig:rq4}, HyperAgent significantly outperforms In-N-Out and semantic retrieval. This result indicates that, with schema-node anchoring and dependency-guided expansion, the extracted tool context recovers implicit prerequisite tools that are not directly aligned with the task semantics, thereby providing a reliable planning and execution space for task completion.

\myparagraph{RQ5: Can HyperAgent Propose Better Tool-Use Sequences under Dynamical States}
\begin{figure}[t]
  \centering
\includegraphics[width=.55\linewidth]{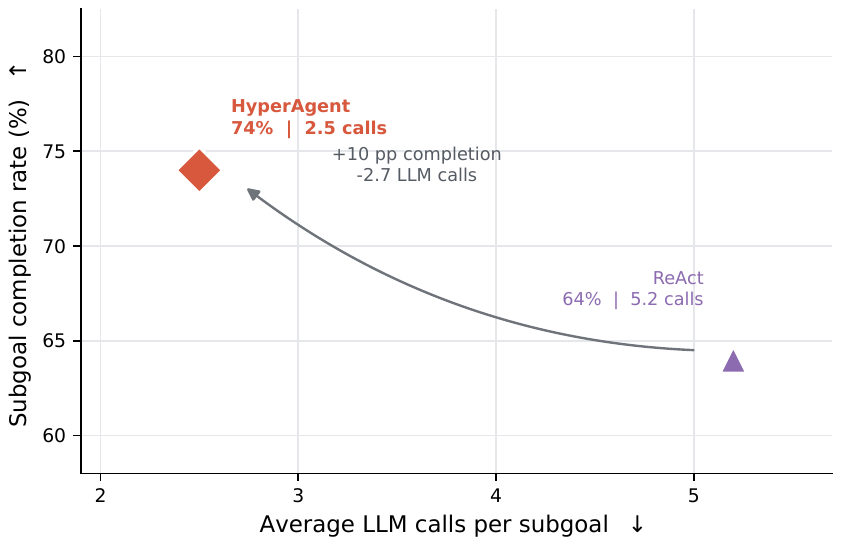}
\caption{Subgoal completion rate under dynamic states.}
  \label{fig:rq5}
\end{figure}
We evaluate the constructed tool support graphs by examining whether they improve subgoal completion while reducing LLM calls. 
We compare HyperAgent with ReAct using semantic Top--K retrieval under the same tool budget. 
For a fair comparison, both methods operate on the same frozen set of subgoals. 
Before executing either method, we manually annotate completion contracts for a predefined subset of subgoals, specifying their expected effects and outputs. 
A subgoal is considered completed only when its contract is supported by runtime observations or verified changes to the environment.
Based on the results in Figure~\ref{fig:rq5}, HyperAgent achieves a higher subgoal completion rate while reducing the average number of LLM calls per subgoal. This result suggests that the state-conditioned support graphs expose prerequisite tools and intermediate schemas before execution, thereby reducing trial-and-error tool selection and repeated exploration relative to the baseline.
\section{Conclusion}
In this paper, we present HyperAgent, a Tool-Schema Hypergraph-guided framework for dynamic tool-use planning with LLM agents. 
We extract schema-level dependencies among tools and represent the available tool space as a directed Tool-Schema Hypergraph. 
Given a task, HyperAgent retrieves a task-relevant Tool Context Graph that preserves schema-level dependencies and guides the decomposition of the task into a fine-grained Task DAG. 
During execution, we further introduce deficit-based support graph expansion to construct a state-conditioned and structurally grounded tool composition for each subtask. 
Experiments on AppWorld demonstrate that HyperAgent improves task completion while reducing redundant API calls, LLM interactions, and token consumption compared with the selected baselines. 
These findings suggest that explicitly modeling tool dependencies can complement the semantic reasoning capabilities of LLMs, enabling more reliable and efficient tool-use planning under dynamically changing agent states.
\clearpage

\bibliography{main}
\end{document}